\documentclass{article} 
\usepackage{nips13submit_e,times}
\usepackage{hyperref}
\usepackage{url}
\usepackage[compact]{titlesec}
\usepackage{amsmath,amsfonts,amssymb,amsthm}
\usepackage{mathtools}
\usepackage{commath}
\usepackage[sc,osf]{mathpazo}
\usepackage{comment}
\usepackage[noadjust]{cite}
\usepackage{graphicx}
\usepackage{caption}
\usepackage{subcaption}
\usepackage{amsmath,amssymb}
\DeclareMathOperator{\E}{\mathbb{E}}

\title{Language Modeling with Generative Adversarial Networks}

\author{
Mehrad Moradshahi
\\
Department of Electrical Engineering\\
Stanford University\\
\texttt{mehrad@stanford.edu} \\
\And
Utkarsh Contractor \\
Department of Computer Science \\
Stanford University \\
\texttt{utkarshc@stanford.edu} \\
}

\nipsfinalcopy 

\begin{document}

\maketitle

\begin{abstract}
Generative Adversarial Networks (GANs) have been promising in the field of image generation, however, they have been hard to train for language generation. GANs were originally designed to output differentiable values, so discrete language generation is challenging for them which causes high levels of instability in training GANs. Consequently, past work has resorted to pre-training with maximum-likelihood or training GANs without pretraining\cite{Press2017ganpretrain} with a WGAN\cite{wassensteingan} objective with a gradient penalty. In this study, we present a comparison of those approaches. Furthermore, we present the results of some experiments that indicate better training and convergence of Wasserstein GANs (WGANs) when a weaker regularization term is enforcing the Lipschitz constraint. 
\end{abstract}

\section{Introduction}
Generative Adversarial Networks (GANs) \cite{goodfellowgan} are used to train generative models in an adversarial setup, with a generator generating images that are trying to fool a discriminator whose role is to discriminate between real and synthetic images. GANs have shown good results and gained much attention when it comes to producing new images, however, language generation is an active and challenging area of research with GANs, mostly due to the non-differentiable nature of generating discrete symbols. Although GANs can accurately model complex distributions, they are known to be difficult to train due to instabilities caused by a difficult minimax optimization problem in a game theoretic situation, where the generator and discriminator aim to find a Nash Equilibrium. 

Formally, the game between the generator G and the discriminator D is the minimax objective:


\begin{equation} \label{equ1}
  min_G max_D \E_{x \sim\ P_r}[log(D(x))] + \E_{\hat{x} \sim\ P_g}[log(1 - D(\hat{x}))]  
\end{equation}

where $Pr$ is the data distribution and $Pg$ is the model distribution implicitly defined by $\hat{x} = G(z), z = p(z)$ (the input $z$ to the generator is sampled from some simple noise distribution p, such as the uniform distribution or a spherical Gaussian distribution). Wasserstein GANs (WGANs) with gradient penalty provide stable training of GANs and provide a better approach to reasonably train GANs. \cite{Gulrajaniwgan2017}

In this paper, we will be evaluating the use of Generative Adversarial Networks (GANs) for text generation and language modeling. In particular we look at two implementations \cite{Press2017ganpretrain} and \cite{sai2017advgan} using WGAN with gradient penalty. Furthermore, we provide some experimental results by training GANs with weaker regularization\cite{reggan} and a modified gradient penalty that provides indications of better training stability of WGANs.

\begin{figure}[!h]
\centering
\includegraphics[width=0.6\textwidth]{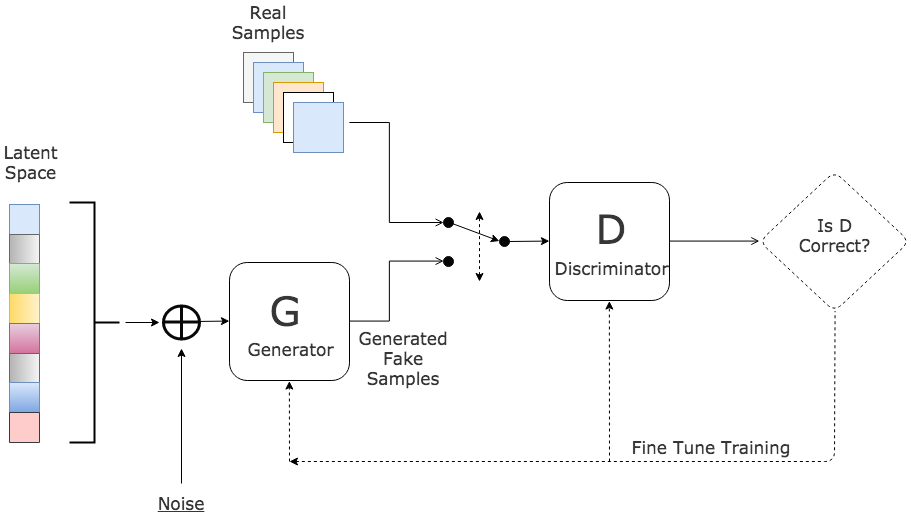}
\caption{GAN general architecture}
\label{NN}
\end{figure}

\section{Related work}
Sai et. al \cite{sai2017advgan} have introduced a simple baseline that addresses the discrete output space problem without relying on gradient estimators and shows that it is able to achieve state-of-the-art results on a Chinese poem generation dataset and presented quantitative results on generating sentences from context-free and probabilistic context-free grammars, and qualitative language modeling results. A conditional version is also described that can generate sequences conditioned on sentence characteristics.

Ofir et. al \cite{Press2017ganpretrain} have shown that recurrent neural networks can be trained to generate text with GANs from scratch using curriculum learning, by slowly teaching the model to generate sequences of increasing and variable length. They empirically show that their approach vastly improves the quality of generated sequences compared to a convolutional baseline.

Henning et. al \cite{reggan} present theoretical arguments why using a weaker regularization term enforcing the Lipschitz constraint is preferable. These arguments are supported by experimental results on several data sets. For stable training of Wasserstein GANs, they propose to use the following penalty term to enforce the Lipschitz constraint that appears in the objective function:


\begin{equation} \label{equ2}
  \E_{\hat{x} \sim\ \tau}[(max\{0,\hspace{1mm}\norm{\nabla f(\hat{x})}_2 - 1\})^2]  
\end{equation}

\section{Data}
We have used the One Billion Word Benchmark for Measuring Progress in Statistical Language Modeling\cite{Cipriandata2013}. The project makes available a standard corpus of 0.8 billion words to train and evaluate language models. Duplicate sentences were removed, dropping the number of words from about 2.9 billion to about 0.8 billion. Words outside of the vocabulary were mapped to [unk] token, also part of the vocabulary. Sentence order was randomized, and the data was split into 100 disjoint partitions One such partition (1\%) of the data was chosen as the held-out set. The benchmark is available as a code.google.com project at \url{https://code.google.com/p/1-billion-word-language-modeling-benchmark/}; besides the scripts needed to rebuild the training/held-out data, it also makes available log-probability values for each word in each of ten held-out data sets, for each of the baseline n-gram models.

\section{Methods} \label{methods}
We have used the implementation in \cite{Press2017ganpretrain} and \cite{sai2017advgan} as our baseline. In both cases, we have used the 1 billion word dataset. GAN based methods have often been critiqued for lacking a concrete evaluation strategy \cite{improvedgantrain} and resort to some sort or subjective evaluation. For our evaluations, in \cite{Press2017ganpretrain} we generate 640 sequences from each model and measure \%-IN-TEST-n, that is, the proportion of word n-grams from generated sequences that also appear in a held-out test set. We evaluate these metrics for n ∈ {1, 2, 3, 4}., however for \cite{sai2017advgan} we just rely on human understanding and thus is qualitative. 

In both cases, we have used multiple hyper-parameter variations to find the most optimal set. For \cite{Press2017ganpretrain}, we were able to improve the results by increasing the batch-size and increasing the number of epochs. Recently, Wasserstein GAN (WGAN) have been introduced which solve the vanishing gradient problem by clipping the weight values to lie in a cube. However, it has been shown that the gradient signals that generator receives might become unstable. An improved version uses weaker regularization for gradient penalty instead of clipping to force that double-sided gradient approaches. We have implemented this method and used it with a model trained based on \cite{Press2017ganpretrain}. Training duration for GANs is unreasonably long, considering its reaches convergence at all. For \cite{Press2017ganpretrain} we have trained the network on four K80 GPUs for up to 14 days to run the variable length curriculum learning with teacher helping to generated sequences of length up to 25 with multiple hyper-parameters selections.

\begin{figure}[!h]
\centering
\includegraphics[width=1\textwidth]{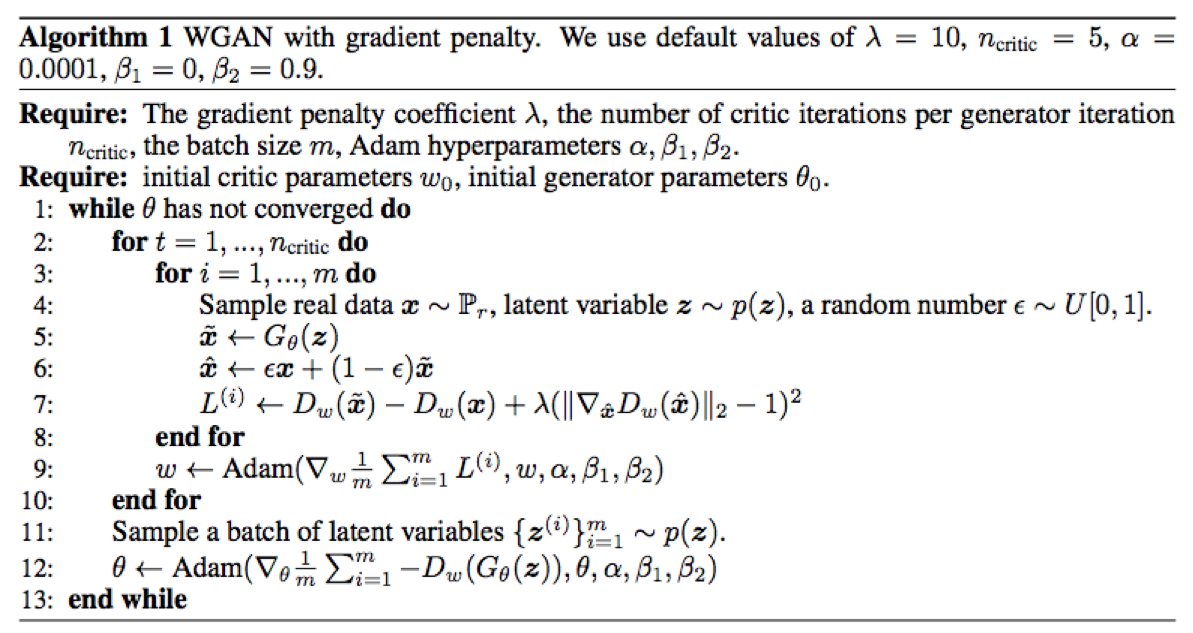}
\caption{Improved WGAN algorithm}
\label{NN}
\end{figure}

\section{Experiments/ Results/ Discussion}
We used Amazon Elastic Compute Cloud and Microsoft Azure virtual machines to run our experiments. They provided the necessary CPU and GPU power to scale up our algorithms and run parallel processes. We evaluated our first model by generating sequences from the model and measuring "percent in Test-N", i.e. the proportion of word n-grams from generated sequences that also appear in a held-out test set. The goal is to measure the extent to which the generator is able to generate real words with local coherence. In the second model, we use human perception to decide on the quality of the produced text. 

\subsection{Language Generation with Recurrent Generative Adversarial Networks without Pre-training}
This section follows the implementation on the lines of \cite{Press2017ganpretrain} where we train GRU based RNNs \cite{arxiv2014rnngru} to generate text using WGANs from scratch using curriculum learning, by teaching the model to generate sequences of increasing and variable length on the Billion word dataset \cite{Cipriandata2013}. We trained our model for up to 25 sequences on variable batch sizes, epochs, network layers, sequence lengths and observed the following results (Table \ref{tab:exp1})

\begin{table}[h!]
\begin{center}

\begin{tabular}{*{5}{|c}|}
\hline
Generated Samples & Unigrams & Bigrams & Trigrams & Quadgrams \\
\hline \hline
"I have more than the man was" & & & &\\
                    "The NBC said so , Most of the" & 0.84 & 0.57 & 0.27 & 0.03 \\
                    "Reid Maybe you got the more than" & & & & \\
\hline
\end{tabular}

\end{center}
\caption{Experiment 1 results} \label{tab:exp1}
\end{table}

The proposed idea was to augment the WGAN loss by a regularization term that penalizes the deviation of the gradient norm of the critic with respect to its input from one (leading to a variant referred to as WGAN-GP, where GP stands for gradient penalty).\cite{Press2017ganpretrain}. However, we observed that increasing the batch size to 128 and using a weaker regularization term enforcing the Lipschitz constraint\cite{reggan}, led to highest accuracy across our experiments, and as early as sequence length 11 resulting to a Unigrams and Bigrams \%-in-N accuracy of 85\% and 57\% respectively. 

We also noticed an improvement in the stability and convergence of the GAN training when we used a modified gradient penalty \cite{reggan} with the WGAN objective, as seen in (Figure 4) below. More specifically when we use equation $\E_{\hat{x} \sim\ \tau}[(max\{0,\hspace{1mm}\norm{\nabla f(\hat{x})}_2 - 1\})^2]$ to enforce the Lipschitz constraint that appears in the objective function instead of the previously considered approaches of clipping weights and of applying the stronger gradient penalty $ \E_{\hat{x} \sim\ \tau}[(\hspace{1mm}\norm{\nabla f(\hat{x}\hspace{1mm})}_2 - 1)^2]$.

In addition to more stable learning behavior, the proposed regularization term leads to lower sensitivity to the value of the penalty weight $\lambda$ (demonstrating smooth convergence and well-behaved critic scores throughout the whole training process for different values of $\lambda$)

\begin{figure}[!h]
  \centering
  \begin{minipage}[b]{0.49\textwidth}
    \includegraphics[width=\textwidth]{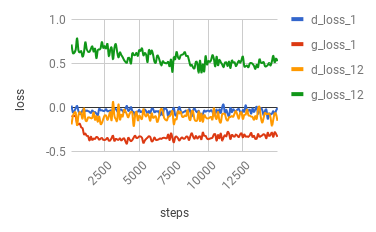}
    \caption{WGAN with original GP}
  \end{minipage}
  \hfill
  \begin{minipage}[b]{0.49\textwidth}
    \includegraphics[width=\textwidth]{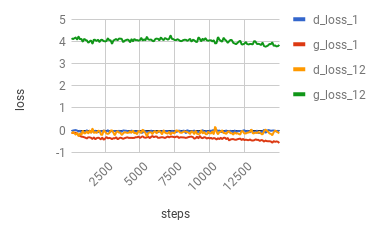}
    \caption{WGAN with modified GP}
  \end{minipage}
\end{figure}

\subsection{Adversarial Generation of Natural Language}
This section follows the implementation on the lines of \cite{sai2017advgan}. We have access to a variety of generator and discriminator architectures including LSTM based RNN and 1D CNN. So far we got the best results by using LSTM for both generator and discriminator. We trained our model for different objective functions to generate 64 batches of 5-word sequences after every 50 iterations. As you see in Table \ref{tab:exp2}, WGAN outperform GAN, and WGAN with GP outperforms WGAN.\\ We used both quantitative and qualitative measures to compare the results. As you see GAN makes frequent grammatical mistakes and repeats same words in a phrase. We see these errors a lot in the earlier stages of training, but even after many iterations, GAN keeps making those mistakes. WGAN has two advantages. It trains faster than GAN and produces more coherent results. Even though it still produce illegible phrases occasionally, it learns much faster not to repeat the same mistakes again. Adding gradient penalty makes WGAN more robust to errors and training instability. If not being set properly, the training process can be highly unstable and the loss starts bouncing around. We Used cross-validation on dev set to tune the hyper-parameters. Considering it takes a day to run each experiment, we used a small set of candidate values for hyper-parameters and used cross-validation on dev set to tune them. \\
The score indicates the fraction of all sentences produced during the training, which are not in the corpus. It is a rough estimate of how well the generator is performing.\\
After running the code for 5 epoch we got the following results:

\begin{table}[h!]
\begin{center}

\begin{tabular}{*{5}{|c}|}
\hline
Objective function & Samples & Score \\
\hline \hline
 & "he only [unk] hard ?" & \\
GAN & "about about in month ?" & 0.85 \\
 & "the are are murder !"  & \\
\hline
& "like last doing monday arrested" & \\
      WGAN   & "to all businesses won ..." & 0.88 \\
      & "they have were arrested ?" & \\
      \hline
      & "your even example killed it" & \\
    WGAN with GP  & "he could have started ." & 0.90 \\
    & "finally cannot do parties ?" & \\
\hline
\end{tabular}

\end{center}
\caption{Experiment 2 results} \label{tab:exp2}
\end{table}

\section{Conclusion/ Future Work}
In conclusion, we show that GANs can produce original text with local coherency. Use of WGAN\cite{wassensteingan} is preferred over GAN as it trains faster, with higher stability, and produces more coherent phrases. Furthermore, we also show that using a weaker regularization term enforcing the Lipschitz constraint in the WGAN gradient penalty, has promised for more stable GAN training and smoother convergence.
\\
In future work, we would like to explore more algorithms and techniques such as MaskGAN
\cite{maskgan}, where we would like to extend the results of our experiments and evaluate its performance on MASKGAN which introduces an actor-critic conditional GAN that fills in missing text conditioned on the surrounding context. We would also like to train GANs on more coherent text corpus such as the Wikipedia dataset to generate language that is more globally coherent.  As previously mentioned, GANs are challenging when it comes to training stability and convergence and the hyper-parameters are sensitive to small changes; With more time and computation power, we can run these algorithms for longer time and more variations.

\section*{Acknowledgments}

We would like to thank Sandeep Subramanian\cite{sai2017advgan} and Sai Rajeswar\cite{sai2017advgan} for their help during the course of this paper.


\end{document}